\documentclass[runningheads]{llncs}
\usepackage[T1]{fontenc}
\usepackage{graphicx}
\usepackage{amsmath}
\usepackage{amsfonts}
\usepackage{float}
\usepackage{multirow}
\usepackage{xcolor}
\usepackage{appendix}
\usepackage{ulem}
\usepackage{hyperref}
\usepackage{stmaryrd}
\usepackage{color}

\begin{document}
\title{XRFormer: Multiscale Tokenization for XRF Representation Learning}
%
%
\author{Sofiane Daimellah\inst{1} \and
Sylvie Le Hégarat-Mascle\inst{1} \and
Clotilde Boust\inst{2}}
\authorrunning{S. Daimellah et al.}
%
\institute{Université Paris-Saclay, Gif-sur-Yvette, France \and
Centre de Recherche et de Restauration des Musées de France, Paris, France\\
\email{email1@universite-paris-saclay.fr, email2@culture.gouv.fr}}
\maketitle              
\begin{abstract}
X-ray fluorescence (XRF) spectroscopy is a key modality for material analysis in cultural heritage. However, automated learning from XRF spectra remains challenging: XRF spectra are complex one-dimensional signals composed of sharp elemental peaks, broader structures, and background variations that are not taken into account by existing learning-based models. This paper introduces XRFormer, a transformer architecture tailored to XRF spectra through a multiscale convolutional tokenizer that injects locality and multi-resolution inductive biases before global self-attention. The tokenizer progressively reduces spectral resolution while increasing embedding dimensionality, and the resulting token sequence is processed by a standard transformer encoder. We further investigate self-supervised pretraining for XRF representation learning using Masked Spectral Modeling (MSM) and a physics-informed Peak Presence Prediction (PPP) objective. Experiments on the Pigments Checker STANDARD v.5 dataset for pigment identification and unmixing show that XRFormer consistently outperforms ViT, SpectralFormer (with and without CAF), and a 1D-CNN baseline for pigment identification. For pigment unmixing, XRFormer achieves robust abundance estimation while maintaining significantly higher parameter efficiency than SpectralFormer, operating at a lower token resolution (128 vs. 512 tokens) and with less than half the number of parameters (1.5M vs. 3.37M). MSM yields consistent gains across both tasks, while PPP further enhances performance for both identification and unmixing when tuned with an appropriate peak prominence. These results highlight multiscale, modality-aware tokenization as an effective and parameter-efficient foundation for transformer-based XRF modeling under data-limited conditions. A GitHub repository is provided at \url{https://github.com/sofiane1010/XRFormer}.

\end{abstract}
\section{Introduction}
\begin{figure}[ht]
    \centering
    \includegraphics[width=0.75\linewidth]{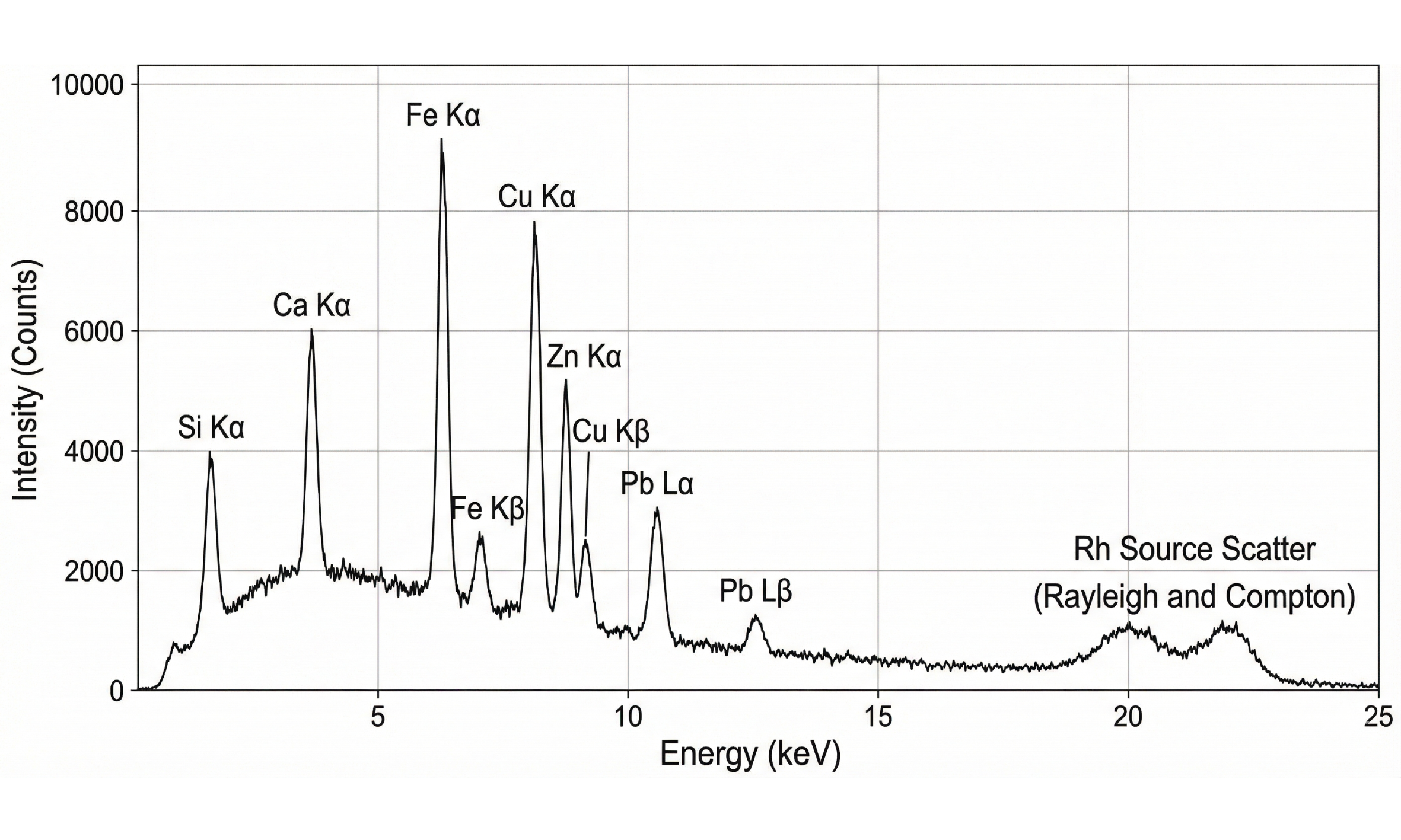}
    \caption{Typical XRF spectrum showing the energies of fluorescent emission lines (peaks) corresponding to different elements.}
    \label{fig:xrf_spectrum}
\end{figure}

X-ray fluorescence (XRF) is a non-destructive technique for elemental analysis that measures characteristic X-ray emissions produced under X-ray excitation~\cite{bezur2020handheld}. Each element emits photons at specific energies, yielding a one-dimensional spectrum over energy channels composed of sharp emission peaks superimposed on a continuous background (Figure~\ref{fig:xrf_spectrum}). XRF is widely used in medical diagnostics~\cite{Börjesson_Mattsson_2007}, environmental analysis~\cite{kalnicky2001field}, and geology and mining~\cite{oyedotun2018x}, and it plays a central role in Cultural Heritage (CH) by enabling non-invasive characterization of artistic materials such as pigments and metals~\cite{ravaud2016development}.

Pigment identification and unmixing are common CH applications of XRF, yet automated analysis remains limited. Most existing studies focus on pigment identification and often rely on expert-driven peak interpretation. Recent learning-based approaches operate directly on raw spectra using one-Dimensional Convolutional Neural Networks (1D-CNNs)~\cite{jones2022neural,xu2022can}. While 1D-CNNs effectively capture local patterns such as elemental peaks, they struggle at modeling long-range relationships across the energy axis. Such global dependencies arise from correlated elemental emissions and background effects and are not explicitly captured by locally constrained architectures. Transformers provide a natural mechanism to model these long-range interactions via self-attention.

Transformers were introduced for sequence modeling~\cite{vaswani2017attention} and were later adapted to images by the Vision Transformer (ViT)~\cite{dosovitskiy2020image}. While a recent work has explored the application of transformers to XRF datacubes for virtual painting recolouring~\cite{bombini2025towards}, architectures explicitly tailored to the fine-grained spectral structure and physical characteristics of 1D XRF signals remain largely unexplored. Related progress in hyperspectral imaging (HSI), which also relies on one-dimensional spectral signals, shows that adapting the tokenization stage is critical. SpectralFormer~\cite{hong2021spectralformer} introduced a group-wise spectral embedding to incorporate local continuity, while MCE-ST~\cite{khotimah2023mce} employed convolutional spectral embeddings to enhance local feature extraction. However, XRF spectra contain both narrow peaks and broader spectral structures, motivating representations that combine locality with multiscale context. Inspired by the Multiscale Vision Transformer (MViT)~\cite{fan2021multiscale}, we propose a multiscale convolutional tokenizer that progressively reduces spectral resolution while increasing embedding dimensionality. The resulting multiscale token sequence is processed by a standard transformer encoder, enabling global attention over representations that already encode multi-resolution spectral information.

We also investigate self-supervised learning (SSL) to improve representation quality under limited labeled data. In spectral transformers, MAEST~\cite{ibanez2022masked} extended SpectralFormer with a masked modeling objective and reported gains on hyperspectral classification. Building on this, we assess SSL pretraining for XRF transformers and quantify its impact on pigment identification and unmixing.

In summary, this work makes the following contributions. First, we introduce a multiscale convolutional tokenizer tailored to XRF spectra, which captures both narrow elemental peaks and broader spectral structures before transformer-based processing. Second, we study self-supervised pretraining strategies for XRF spectral modeling, considering both a general Masked Spectral Modeling (MSM) objective and a physics-informed Peak Presence Prediction (PPP) task. PPP is defined using peak cues extracted directly from the input spectrum and does not rely on pigment labels, providing a modality-specific pretext task. Finally, we evaluate the proposed architecture, with and without SSL pretraining, on pigment identification and compare it against ViT, SpectralFormer and 1D-CNN models on pigment identification and pigment unmixing.

\section{Related Work}
\subsection{Vision Transformer Adaptations for Spectral Modeling}\label{sec:rel_work_vit}

Transformers were originally introduced for sequence modeling~\cite{vaswani2017attention} and later adapted to vision through architectures such as the Vision Transformer (ViT)~\cite{dosovitskiy2020image} and subsequent variants~\cite{fan2021multiscale,liu2021swin}. 
For hyperspectral signals, several works adapt ViT-style models by revisiting tokenization and spectral embedding to accommodate spectral data~\cite{hong2021spectralformer,liu2022mapping,zhang2022convolution,sun2022spectral,he2022csit,khotimah2023mce}. SpectralFormer, for instance, introduces a group-wise spectral embedding that builds overlapping tokens from adjacent spectral bands before transformer encoding~\cite{hong2021spectralformer}. Other approaches incorporate convolutional operations into the embedding process: CTMixer enriches tokens with local spectral context, while SSFTT combines spectral dimensionality reduction with convolutional processing and a Gaussian-weighted tokenizer~\cite{zhang2022convolution,sun2022spectral}. Alternatively, CSiT proposes a multiscale spectral embedding strategy that processes spectral bands at multiple resolutions and exchanges information across scales via cross-attention~\cite{he2022csit}.

Overall, prior spectral transformers mainly emphasize either local spectral continuity (via convolutional or overlapping embeddings) or multiscale processing. However, XRF spectra contain both sharp, localized peaks and broader background structures, which require token representations that encode locality while preserving multiscale context. The multiscale convolutional tokenizer proposed in this work is designed to meet this requirement.

\subsection{Learning-Based Approaches for XRF Pigment Identification}\label{sec:rel_work_pigment_id}
Relatively few works have addressed pigment identification from XRF spectra using data-driven approaches. A representative method applies unsupervised clustering to a macro-XRF datacube to group pixels with similar spectral signatures; pigment identification is then performed by manually inspecting characteristic energy peaks in the cluster-averaged spectra and assigning pigments based on elemental composition~\cite{kogou2021new}. More recent studies explore supervised models based on 1D-CNNs trained directly on raw XRF spectra~\cite{jones2022neural,xu2022can}. 
However, existing XRF-based methods remain either semi-automatic or predominantly local in their spectral processing, which can limit their ability to capture global dependencies across distant energy regions. In contrast, the transformer-based model proposed in this work enables modeling long-range relationships across the full XRF spectrum through self-attention.

\subsection{Self Supervised Learning}\label{sec:rel_work_ssl}
Self-supervised pretraining learns transferable representations by exploiting intrinsic data structure, typically through two complementary paradigms~\cite{xu2023multimodal}. First, \emph{application-agnostic} objectives, such as masked modeling~\cite{devlin2019bert} or contrastive learning~\cite{radford2021learning}, focus on capturing general structural patterns in the input modality. 
Second, \emph{application-specific} objectives incorporate domain knowledge to address specialized needs where generic pretraining is insufficient, as demonstrated in domains like source code~\cite{guo2020graphcodebert} and medical records~\cite{rasmy2021med}. In practice, tailored pretexts often complement generic objectives by injecting inductive biases aligned with the target application.

Building on these two paradigms, we investigate an application-agnostic MSM objective, alongside an application-specific, physics-informed PPP task, and quantify their impact on downstream XRF-based pigment identification and unmixing.

\section{Adapting Vision Transformers to XRF Spectra}
\subsection{Tokenization and Embedding Space}\label{sec:tokenization}

\begin{figure}[ht]
    \centering
    \includegraphics[width=0.85\textwidth]{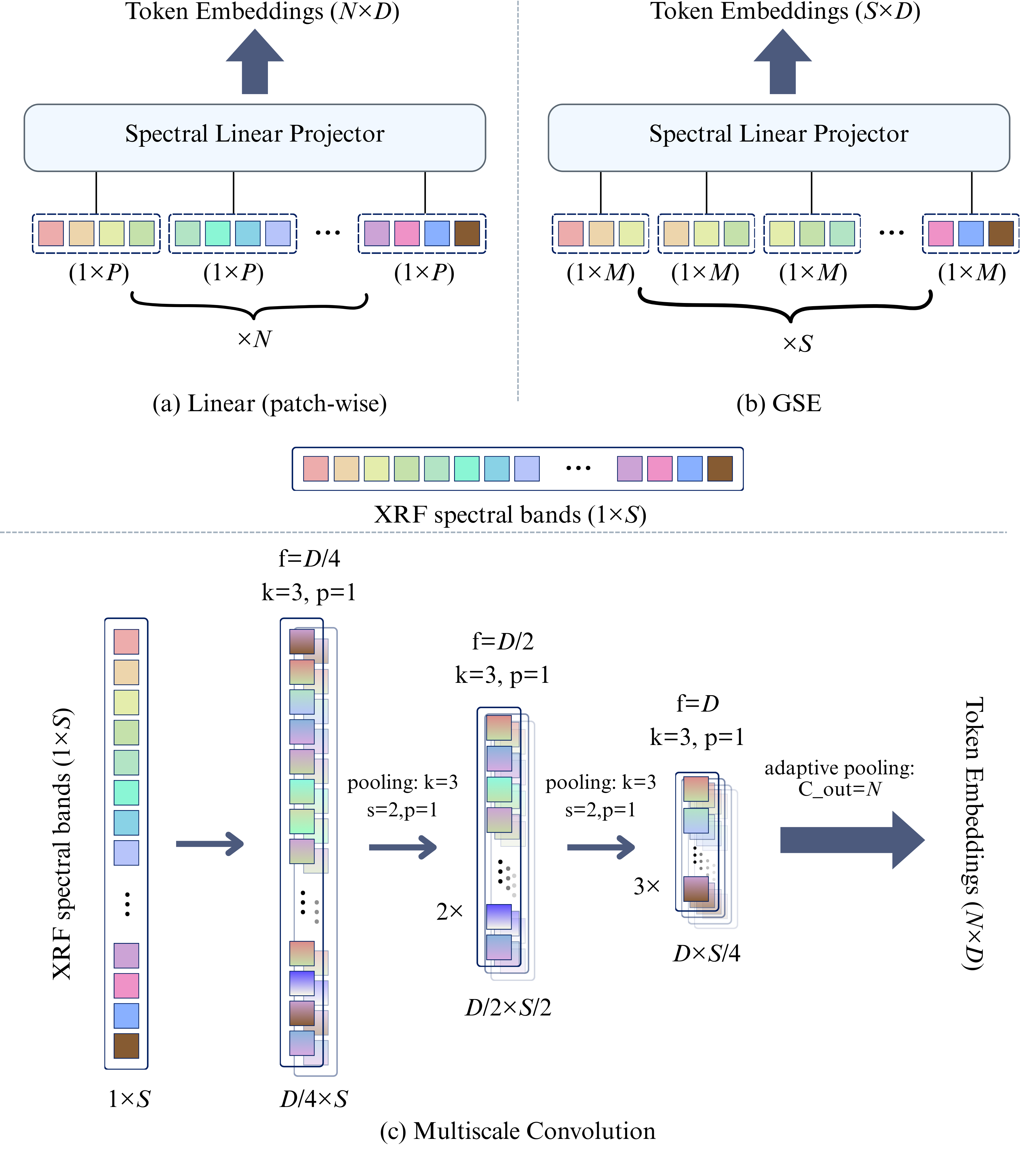}
    \caption{Three spectral tokenization strategies: (a)~linear patch-wise embedding (ViT), (b)~group-wise spectral embedding (SpectralFormer), and (c)~multiscale convolutional embedding. All tokenizers map a raw XRF spectrum to a token sequence processed by the same transformer encoder.}
    \label{fig:tokenizers}
\end{figure}

We consider three representative tokenization strategies for adapting transformer architectures to 1D XRF spectra: (a)~linear (patch-wise) embedding (ViT~\cite{dosovitskiy2020image}), (b)~group-wise spectral embedding (SpectralFormer~\cite{hong2021spectralformer}), and (c)~the multiscale convolution tokenizer proposed in this work (Figure~\ref{fig:tokenizers}).

Given a spectrum $\mathbf{x}\in\mathbb{R}^{1\times S},$ tokenization maps it to a sequence of $N$ tokens (excluding the $[CLS]$ token which will be introduced later) represented in a $D$-dimensional embedding space:  
\begin{equation*}
    \Phi: \mathbb{R}^{1\times S} \rightarrow \mathbb{R}^{N\times D}, \qquad \mathbf{Z}=\Phi(\mathbf{x}).
\end{equation*}

\subsubsection{Linear (patch-wise) Embedding}
Following ViT~\cite{dosovitskiy2020image}, $\mathbf{x}$ is split into $N$ non-overlapping patches of length $P$, forming $\mathbf{x}_P\in\mathbb{R}^{N\times P}$ (Figure~\ref{fig:tokenizers}.a). Each patch is then mapped to a $D$-dimensional space through a learnable linear projection:
\begin{equation*}
    \mathbf{Z} = \mathbf{x}_P \mathbf{W} + \mathbf{b}, \quad
    \mathbf{W} \in \mathbb{R}^{P \times D},\ \mathbf{b}\in\mathbb{R}^{D}.
\end{equation*}

\subsubsection{Group-wise Spectral Embedding (GSE)}
Instead of non-overlapping patches, GSE extracts overlapping windows of size $M$ with a stride of $1$, yielding $S$ patches (special case where $N=S$) $\mathbf{x}'_P\in\mathbb{R}^{S\times M}$ (Figure~\ref{fig:tokenizers}.b). Each patch is then mapped to a $D$-dimensional space through a trainable linear projection $W'$:
\begin{equation*}
    \mathbf{Z} = \mathbf{x}'_P \mathbf{W}' + \mathbf{b}', \quad
    \mathbf{W}'\in\mathbb{R}^{M\times D},\ \mathbf{b}'\in\mathbb{R}^{D}.
\end{equation*}

\subsubsection{Multiscale Convolution}
Our tokenizer progressively increases channel capacity while reducing spectral resolution in order to capture spectral features at multiple scales. Concretely, the input spectrum is processed by three 1D convolutional blocks with increasing effective receptive fields (Figure~\ref{fig:tokenizers}.c). We use fixed kernel sizes and increase the effective receptive field by stacking more convolutional layers. Each convolutional layer is followed by a GeLU activation and batch normalization. At the end of the first and second blocks, strided convolutions are used to downsample the spectral resolution by a factor of two. The final block is followed by an adaptive pooling layer to produce a fixed number of $N$ tokens, yielding $\mathbf{Z}\in\mathbb{R}^{N\times D}$.

Overall, these strategies introduce increasing inductive biases for spectral modeling. Linear patch-wise embedding does not explicitly encode any inductive bias on relative spectral positions within a patch. GSE aggregates local neighborhood at a single scale. In contrast, the proposed tokenizer combines locality with multi-resolution analysis by progressively enlarging the effective receptive field while reducing spectral resolution. 
This design is particularly well aligned with the physical characteristics of XRF spectra, which exhibit narrow elemental peaks, broader peak structures, and slowly varying backgrounds~\cite{bezur2020handheld}.

\subsection{XRFormer Architecture}\label{sec:xrformer}

\begin{figure}[ht]
    \centering
    \includegraphics[width=\linewidth]{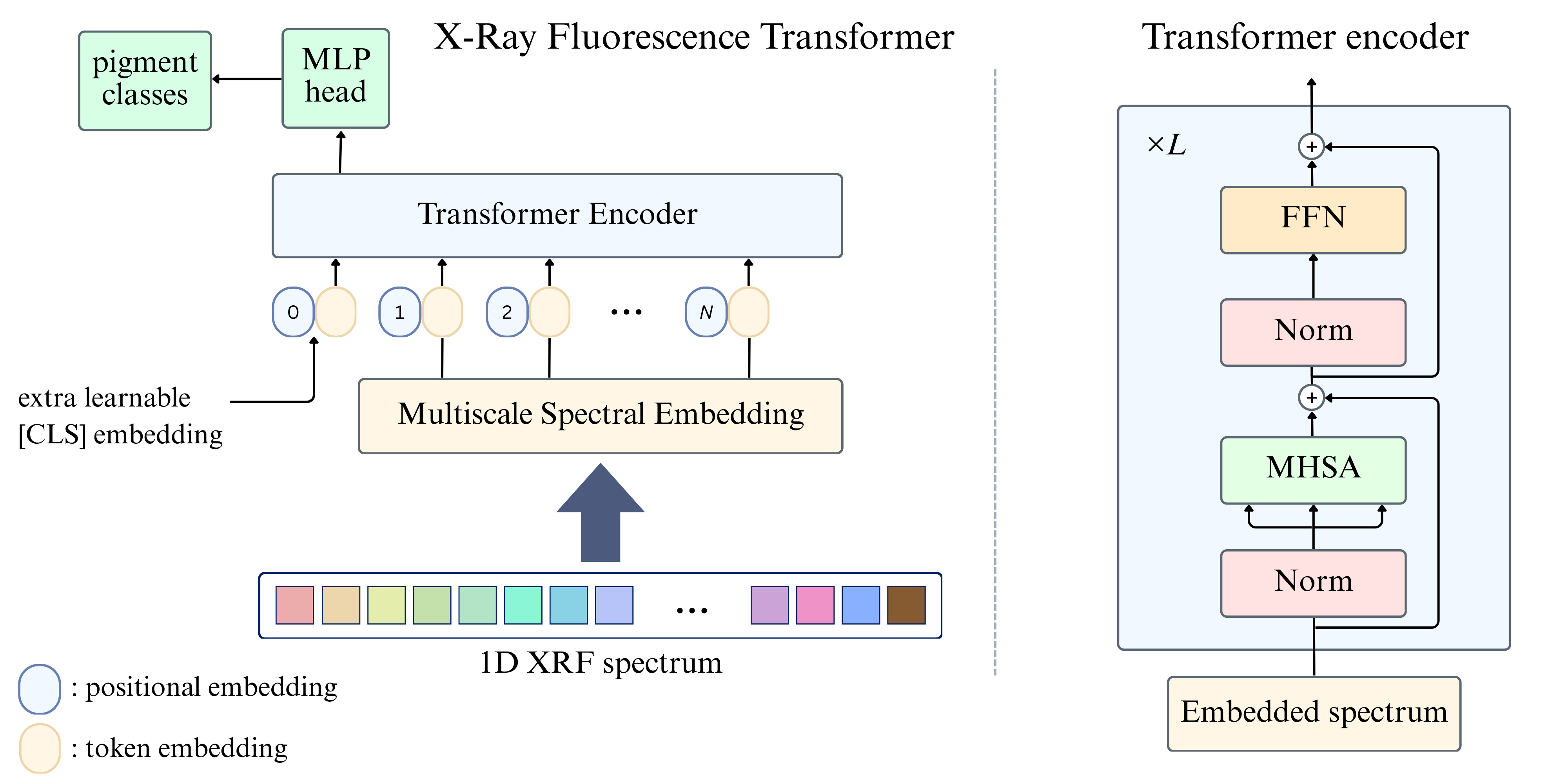}
    \caption{XRFormer overview: the multiscale convolutional tokenizer maps the input XRF spectrum to a sequence of token embeddings, which are processed by a standard transformer encoder. The final $[CLS]$ token serves as a global representation for downstream tasks.}
    \label{fig:xrformer}
\end{figure}

Based on the tokenization analysis above, we adopt the proposed multiscale convolutional tokenizer as the front-end of our model. We now present \emph{XRFormer}, which combines this modality-aware tokenizer with a standard transformer encoder for global spectral modeling (Figure~\ref{fig:xrformer}).

Given an input XRF spectrum $\mathbf{x}\in\mathbb{R}^{1\times S}$, the tokenizer outputs $\mathbf{Z}\in\mathbb{R}^{N\times D}$ which is fed into the transformer encoder. We prepend a learnable $[CLS]$ token, which serves as a global representation of the spectrum, and add learnable positional embeddings to all tokens, including the $[CLS]$ token, to preserve spectral ordering information.

The resulting sequence is processed by a stack of $L$ transformer blocks following the standard pre-normalization formulation~\cite{vaswani2017attention}. Each block consists of a Multi-Head Self-Attention (MHSA) layer and a position-wise Feed-Forward Network (FFN), both having residual connections:
\begin{align}
    \mathbf{Z}'_l &= \mathrm{MHSA}\big(\mathrm{LN}(\mathbf{Z}_l)\big) + \mathbf{Z}_l, 
    \label{eq:mhsa}\\
    \mathbf{Z}_{l+1} &= \mathrm{FFN}\big(\mathrm{LN}(\mathbf{Z}'_l)\big) + \mathbf{Z}'_l, 
    \label{eq:mlp}
\end{align}
where $\mathrm{LN}(\cdot)$ denotes Layer Normalization. MHSA enables global modeling of long-range spectral dependencies beyond local convolutional receptive fields.

For supervised discriminative tasks such as pigment identification, we apply a task-specific feed-forward head to the final hidden state of the $[CLS]$ token to produce a spectrum-level prediction.

\subsection{Self-Supervised Pretraining Objectives}\label{sec:ssl}

While XRFormer can be trained end-to-end in a supervised manner, we further investigate whether self-supervised pretraining improves the representation quality under limited labeled data, as often encountered in CH applications. We explore two self-supervised objectives that leverage the structure and physical characteristics of XRF spectra: MSM and PPP.

\subsubsection{Masked Spectral Modeling}
Following masked modeling approaches in natural language processing~\cite{devlin2019bert}, we adopt a reconstruction task adapted to one-dimensional XRF spectra. Given a tokenized spectrum, a fixed proportion ($15\%$) of tokens is randomly selected for masking. Among the selected tokens, $80\%$ are replaced by a learnable $[MASK]$ token, $10\%$ are replaced by tokens sampled at random, and the remaining $10\%$ are left unchanged. The model is trained to reconstruct the original spectral values of the masked tokens using a mean squared error loss. By training the model to predict missing tokens from the remaining context, MSM encourages it to capture both local spectral structure (e.g., peak shapes) and long-range dependencies across distant energy bands, which are characteristic of XRF spectra.

\subsubsection{Peak Presence Prediction}
We introduce a physics-informed pretext task to encourage the $[CLS]$ representation to encode characteristic XRF peak distributions. The objective is to predict a binary peak signature $\mathbf{b} \in \{0, 1\}^N$ from the global $[CLS]$ token using a multi-label linear head $W_{PPP} \in \mathbb{R}^{D\times N}$ and a binary cross-entropy loss. The ground-truth signatures $\mathbf{b}_{gt}$ are generated by identifying emission peaks via local maxima detection constrained by prominence. For all $n\in \llbracket 1, N\rrbracket$, we set $b_n = 1$ if a peak falls within the $n$-th token's range $[\frac{nS}{N}, \frac{(n+1)S}{N}]$, and $0$ otherwise. Tuning the prominence involves a trade-off where higher values tend to prioritize dominant (assumed discriminative) peaks for identification, while lower values tend to capture the subtle features required for precise unmixing.

\section{Experiments}
\subsection{Datasets}\label{sec:datasets}

The \textbf{Pigments Checker STANDARD v.5 (PCSv5)} dataset is a reference collection of $69$ historically important pigments spanning prehistory to the mid-20th century~\cite{chsos_pigments}. Each pigment is represented by a single reference XRF spectrum. In this work, we select a subset of $22$ pigments for pigment identification and unmixing experiments (the list of selected pigments is provided in Section~1 of the supplementary material). For models requiring pretraining, we also use the \textbf{Infraart} dataset~\cite{cortea2023infra}, which comprises $1015$ reference XRF spectra obtained from various CH materials.

All spectra are provided as one-dimensional signals of length $2048$ and are uniformly downsampled to $512$ channels for computational efficiency. Both PCSv5 and Infraart are reference-style datasets in which each pigment or material is described by a single spectrum rather than multiple independently acquired measurements. As a result, conventional train/validation/test splits on distinct real samples are not feasible, and data augmentation is necessary to construct learning and evaluation sets. Accordingly, the Infraart dataset is augmented to approximately two million samples for pretraining, while the downstream PCSv5 subset is augmented to $2000$ samples per experiment.

Augmentation is performed by simulating linear mixtures of reference spectra with noise and intensity perturbations to model XRF acquisition variability. Each augmented sample combines two to three reference spectra using weights drawn from a Dirichlet distribution with uniform concentration. The resulting spectrum is scaled by a random intensity factor uniformly sampled in $\left[0.5, 2.0\right]$ to reflect excitation and detector variability, and Poisson noise is applied by sampling photon counts with rate $\lambda = \mathbf{x} \cdot 10^4$ with $\mathbf{x}$ the mixed spectrum. Linear mixture simulation under these conditions is standard in XRF-based CH studies, since elemental contributions combine approximately linearly under typical acquisition settings~\cite{bezur2020handheld,moreau2023multi}.

\subsection{Model Instantiation}\label{sec:model_inst}
Table~\ref{tab:xrformer_models} summarizes the XRFormer model configurations considered in this work. We instantiate XRFormer at two capacity levels, denoted XRFormer-B and XRFormer-L, which differ in transformer depth, embedding dimensionality, FFN hidden dimension, and number of attention heads, while sharing the same multiscale convolutional tokenizer. Unless stated otherwise, XRFormer-B is used as the reference configuration and is referred to as XRFormer in the remainder of the paper.

XRFormer is compared against ViT, SpectralFormer, and a 1D-CNN baseline. For a fair comparison, ViT and SpectralFormer are instantiated with the same transformer depth, embedding dimension, FFN dimension, and number of attention heads as the corresponding XRFormer configuration, differing only in their tokenization and embedding strategies. All transformer-based models operate on a fixed sequence length of 128 tokens, except SpectralFormer, which preserves the full spectral resolution and processes 512 spectral tokens. To further isolate the impact of the tokenizer design, SpectralFormer is evaluated both with and without the Cross-layer Adaptive Fusion (CAF) block; for details on CAF, we refer the reader to the original SpectralFormer paper~\cite{hong2021spectralformer}.

\begin{table}[t]
\centering
\renewcommand{\arraystretch}{1.6}
\scriptsize
\caption{XRFormer model configurations.}
\label{tab:xrformer_models}
\begin{tabular*}{0.7\linewidth}{@{\extracolsep{\fill}}lcccc}
\hline
Model & Layers & Embedding Dim & FFN Dim & \#Heads \\
\hline
XRFormer-B & 6  & 128 & 512  & 8  \\
XRFormer-L & 8  & 256 & 1024 & 16 \\
\hline
\end{tabular*}
\end{table}

\subsection{Training Configuration}\label{sec:training_config}
All models are trained using the Adam optimizer with a batch size of 128. Learning rates are set to $1\times 10^{-4}$ for pretraining and $5\times 10^{-5}$ for downstream fine-tuning. In both cases, a linear learning-rate decay is applied upon validation plateau. For both datasets, augmented samples are randomly split into training, validation, and test sets. For Infraart pretraining, we use an $80/10/10$ split, while for downstream experiments on the PCSv5 subset, a $60/20/20$ split is adopted.

Two pretrained variants of XRFormer are considered. The first is pretrained using MSM only, and the second combines MSM and PPP, following the formulations introduced in Section~\ref{sec:ssl}. Pretraining is performed for up to 400 epochs with early stopping (patience of 30).

For the downstream tasks, namely pigment identification/unmixing, a classification/regression head is attached to the $[CLS]$ token. Models are fine-tuned (end-to-end) for up to $1000$ epochs with early stopping (patience of $60$), using binary cross-entropy and mean absolute error as loss functions for multilabel pigment identification and pigment unmixing, respectively. 

\subsection{Evaluation Metrics}\label{sec:eval_metrics}

We evaluate the proposed models on two downstream tasks: pigment identification and pigment unmixing. Each task is assessed using metrics commonly adopted in multilabel classification and spectral unmixing, respectively.

\subsubsection{Pigment identification}
Let $\mathbf{y}_i \in \{0,1\}^K$ denote the ground-truth presence vector for the $i$-th sample over $K$ pigments, and $\hat{\mathbf{y}}_i$ the corresponding prediction.

Absolute Accuracy (AA) measures the fraction of samples for which \textit{all} pigment labels are correctly predicted:
\begin{equation}
\mathrm{AA} = \frac{1}{N} \sum_{i=1}^{N} \mathbb{I}\left( \hat{\mathbf{y}}_i = \mathbf{y}_i \right),
\end{equation}
where $N$ is the sample number and $\mathbb{I}(\cdot)$ denotes the indicator function.

Hamming Accuracy (HA) evaluates label-wise correctness averaged over all samples and pigments:
\begin{equation}
\mathrm{HA} = 1 - \frac{1}{NK} \sum_{i=1}^{N} \sum_{k=1}^{K} \mathbb{I}\left( \hat{y}_{i,k} = y_{i,k} \right).
\end{equation}
Macro-F1 is computed by first evaluating the F1-score independently for each pigment and then averaging across pigments, ensuring equal importance for rare and frequent pigments.

\subsubsection{Pigment unmixing}
Let $\mathbf{a}_i \in \mathbb{R}^K$ denote the ground-truth abundance vector for sample $i$, $\hat{\mathbf{a}}_i$ the predicted abundance vector, $\mathbf{x}_i$ the ground-truth spectrum, and $\hat{\mathbf{x}}_i$ the reconstructed spectrum.

Abundance RMSE measures the error in estimated pigment proportions:
\begin{equation}
\text{A-RMSE} =
\sqrt{
\frac{1}{NK} \sum_{i=1}^{N}
\left\| \hat{\mathbf{a}}_i - \mathbf{a}_i \right\|_2^2
}.
\end{equation}

Reconstruction RMSE evaluates the fidelity of the reconstructed spectra:
\begin{equation}
\text{R-RMSE} =
\sqrt{
\frac{1}{NS} \sum_{i=1}^{N}
 \left\| \hat{\mathbf{x}}_i - \mathbf{x}_i \right\|_2^2
},
\end{equation}
where $S$ denotes the number of spectral channels.

Finally, the Spectral Angle Mapper (SAM) measures the angular similarity between reconstructed and ground-truth spectra:
\begin{equation}
\mathrm{SAM} =
\frac{1}{N} \sum_{i=1}^{N}
\arccos \left(
\frac{
\langle \hat{\mathbf{x}}_i, \mathbf{x}_i \rangle
}{
\| \hat{\mathbf{x}}_i \|_2 \, \| \mathbf{x}_i \|_2
}
\right).
\end{equation}

Lower values indicate better performance for RMSE- and SAM-based metrics, while higher values indicate better performance for identification metrics.

\subsection{Results Discussion}\label{sec:res_dis}
Tables~\ref{tab:sota_id}--\ref{tab:scaling_combined} present pigment identification and unmixing results. All metrics are averaged over five independent runs and reported as $mean\pm std$.

Without pretraining, XRFormer achieves competitive or superior performance across both tasks. For pigment identification, it improves AA from $68.76\%$ (ViT) to $71.29\%$ and macro-F1 from $91.01\%$ to $92.54\%$. In the unmixing task, XRFormer outperforms ViT in A-RMSE ($0.0435$ vs. $0.0457$). SpectralFormer achieves the best performance across all unmixing metrics, which is attributed to its higher spectral resolution and larger capacity (512 tokens and 3.37M parameters versus 128 tokens and 1.50M parameters for XRFormer).

MSM pretraining improves XRFormer performance for both tasks, yielding a gain of $4.6\%$ in AA and a reduction of $0.003$ in A-RMSE. Incorporating PPP objective further enhances these results, yielding the highest performance among the XRFormer base configurations. Specifically, the MSM+PPP variant achieves the best identification results across all evaluated models with an AA of $76.78\%$, while also narrowing the gap with SpectralFormer in unmixing by reaching an A-RMSE of 0.0399 and a SAM of 0.1345. Finally, increasing model capacity from XRFormer to XRFormer-L leads to consistent improvements in both tasks. For identification, AA increases from $71.29\%$ to $76.39\%$, while for unmixing, A-RMSE is significantly reduced from $0.0435$ to $0.0339$, indicating that increased model capacity enhances the overall discriminative power and quantification accuracy of the learned representations.

\subsubsection{Impact of the tokenizer design}
Across both downstream tasks, XRFormer demonstrates superior overall utility, particularly in pigment identification, where it outperforms all baselines by a clear margin. In the unmixing task, XRFormer achieves more robust abundance estimation compared to ViT, as evidenced by its lower A-RMSE ($0.0435$ vs. $0.0457$). This is significant because A-RMSE is less sensitive to global intensity scaling and spectral noise, serving as a more direct indicator of pigment quantification quality than reconstruction metrics. While SpectralFormer remains the top performer for unmixing, its advantage is largely driven by its significantly higher spectral resolution and capacity. By operating at a higher token resolution, SpectralFormer is naturally better equipped for extracting and processing fine-grained features that are particularly relevant for spectral unmixing and abundance estimation.

These results indicate that the proposed multiscale convolutional tokenizer effectively captures the key characteristics of XRF signals by combining locality with multi-resolution context before global self-attention. By progressively aggregating information across scales, the tokenizer provides compact yet expressive representations that support both instance-level discrimination and mixture analysis. The strong identification gains, combined with competitive unmixing performance under reduced resolution, suggest that the improvements stem primarily from modality-aligned tokenization rather than increased model complexity.

\subsubsection{Effect of self-supervised pretraining}
Self-supervised pretraining provides a clear advantage, as the combination of MSM and PPP yields the highest scores for the base model on both tasks. MSM helps the model learn the general structure of XRF spectra by reconstructing missing data, which improves both identification and unmixing. Adding PPP further enhances these results by forcing the model to focus on specific emission peaks. However, the effectiveness of PPP depends on a careful choice of peak prominence when creating the ground-truth signatures. As noted in Section~\ref{sec:ssl}, higher prominence values tend to benefit identification by emphasizing the dominant peaks required for instance-level discrimination, whereas lower prominence values favor unmixing by capturing the fine-grained spectral variations necessary for precise abundance estimation. When properly tuned, the combination of MSM and PPP further improves performance on both tasks, reaching an Absolute Accuracy of $76.78\%$ and reducing the A-RMSE to $0.0399$.

It is also worth noting that the combination of pretext tasks can be implemented in multiple ways. For example, UNITER used stochastic task sampling~\cite{chen2020uniter}, while LXMERT used joint optimization with weighted losses~\cite{tan2019lxmert}. While we adopted the former approach in this work, investigating alternative multi-task pretraining schemes for spectral data constitutes a promising direction for future work.

\subsubsection{Effect of model scaling}
Increasing model capacity from XRFormer to XRFormer-L leads to systematic performance improvements across all evaluated metrics and both downstream tasks. For pigment identification, larger capacity yields higher absolute accuracy and macro-F1, indicating improved separability and robustness of the learned representations. For pigment unmixing, scaling similarly improves abundance estimation, reconstruction error, and spectral similarity, with clear reductions in A-RMSE, R-RMSE, and SAM. These results confirm that increased representational capacity benefits both discrimination and mixture analysis.
\begin{table}[ht]
\centering
\renewcommand{\arraystretch}{1.6}
\begin{minipage}{0.9\linewidth}
\scriptsize
\caption{Pigment identification performance on the PCSv5 dataset, averaged over 5 runs and reported as $mean\pm std$. The best and second-best results are shown in \textbf{bold} and \underline{underlined}, respectively. SF: SpectralFormer}
\label{tab:sota_id}
\begin{tabular*}{\linewidth}{@{\extracolsep{\fill}} c | c c c c }
\hline
Model & Pretraining & AA $\uparrow$  & HA $\uparrow$ & F1 $\uparrow$\\
\hline
1D-CNN       & -- & $58.61 \pm 1.67$ & $97.55 \pm 0.10$ & $88.69 \pm 0.39$ \\
SF (w/o CAF) & -- & $64.41 \pm 1.74$ & $97.80 \pm 0.08$ & $89.94 \pm 0.37$ \\
SF           & -- & $61.49 \pm 1.78$ & $97.66 \pm 0.17$ & $89.15 \pm 0.83$ \\
ViT          & -- & $68.76 \pm 2.16$ & $98.03 \pm 0.13$ & $91.01 \pm 0.47$ \\
XRFormer     & -- & $71.29 \pm 1.66$ & $98.35 \pm 0.07$ & $92.54 \pm 0.36$ \\
XRFormer     & MSM & $\underline{75.89 \pm 1.26}$ & $\underline{98.66 \pm 0.06}$ & $\underline{93.99 \pm 0.25}$ \\
XRFormer     & MSM+PPP & $\mathbf{76.78 \pm 1.12}$ & $\mathbf{98.69 \pm 0.05}$ & $\mathbf{94.09 \pm 0.29}$ \\
\hline
\end{tabular*}
\end{minipage}
\end{table}

\begin{table}[ht]
\centering
\renewcommand{\arraystretch}{1.6}
\begin{minipage}{0.9\linewidth}
\scriptsize
\caption{Pigment unmixing performance on the PCSv5 dataset, averaged over 5 runs and reported as $mean\pm std$. The best and second-best results are shown in \textbf{bold} and \underline{underlined}, respectively. SF: SpectralFormer.}
\label{tab:sota_unm}
\begin{tabular*}{\linewidth}{@{\extracolsep{\fill}} c | c c c c }
\hline
Model & Pretraining & A-RMSE $\downarrow$ & R-RMSE $\downarrow$ & SAM (rad) $\downarrow$\\
\hline
1D-CNN       & -- & $0.0550 \pm 0.0019$ & $0.0085 \pm 0.0003$ & $0.2039 \pm 0.0075$ \\
SF (w/o CAF) & -- & $0.0577 \pm 0.0069$ & $0.0070 \pm 0.0009$ & $0.1774 \pm 0.0281$ \\
SF           & -- & $\mathbf{0.0379 \pm 0.0033}$ & $\mathbf{0.0048 \pm 0.0006}$ & $\mathbf{0.1278 \pm 0.0152}$ \\
ViT          & -- & $0.0457 \pm 0.0034$ & $0.0058 \pm 0.0004$ & $0.1443 \pm 0.0165$ \\
XRFormer     & -- & $0.0435 \pm 0.0017$ & $0.0060 \pm 0.0004$ & $0.1587 \pm 0.0106$ \\
XRFormer     & MSM & $0.0405 \pm 0.0034$ & $0.0054 \pm 0.0006$ & $0.1417 \pm 0.0153$ \\
XRFormer     & MSM+PPP & $\underline{0.0399 \pm 0.0047}$ & $\underline{0.0053 \pm 0.0007}$ & $\underline{0.1345 \pm 0.0202}$ \\
\hline
\end{tabular*}
\end{minipage}
\end{table}

\begin{table}[ht]
\renewcommand{\arraystretch}{1.3}
\newcommand{\std}[1]{\scalebox{0.8}{\color{black}{$\pm$#1}}}
\centering
\setlength{\tabcolsep}{2pt} 
\begin{minipage}{0.9\linewidth}

\caption{Impact of model scaling on performance on the PCSv5 dataset. Results are $mean \pm std$.}
\scriptsize

\label{tab:scaling_combined}
\begin{tabular*}{\linewidth}{@{\extracolsep{\fill}} c | c c c | c c c}
\hline
\multirow{2}{*}{Model} & \multicolumn{3}{c|}{Pigment Unmixing} & \multicolumn{3}{c}{Pigment Identification} \\
\cline{2-7}
 & \begin{tabular}[c]{@{}c@{}}A-RMSE $\downarrow$\end{tabular} & \begin{tabular}[c]{@{}c@{}}R-RMSE $\downarrow$\end{tabular} & \begin{tabular}[c]{@{}c@{}}SAM $\downarrow$\end{tabular} & \begin{tabular}[c]{@{}c@{}}AA $\uparrow$\end{tabular} & \begin{tabular}[c]{@{}c@{}}HA $\uparrow$\end{tabular} & \begin{tabular}[c]{@{}c@{}}F1 $\uparrow$\end{tabular} \\
\hline
XRFormer   & 0.0435 & 0.0060 & 0.1587 & 71.29 & 98.35 & 92.54 \\
           & \std{0.0017} & \std{0.0004} & \std{0.0106} & \std{1.66} & \std{0.07} & \std{0.36} \\
\hline
XRFormer-L & $\mathbf{0.0339}$ & $\mathbf{0.0048}$ & $\mathbf{0.1186}$ & $\mathbf{76.39}$ & $\mathbf{98.67}$ & $\mathbf{93.97}$ \\
           & \std{0.0026} & \std{0.0003} & \std{0.0120} & \std{0.91} & \std{0.05} & \std{0.23} \\
\hline
\end{tabular*}
\end{minipage}
\end{table}
\section{Conclusion}
This work presented XRFormer, a transformer-based model designed for one-dimensional XRF spectra with an explicit focus on the tokenization stage. The proposed multiscale convolutional tokenizer introduces locality and multi-resolution processing tailored to the peak-dominated structure and background variability of XRF signals, while retaining a standard transformer encoder for global spectral modeling. Across pigment identification and unmixing on PCSv5, XRFormer consistently outperformed ViT (in abundance estimation for unmixing), SpectralFormer (without CAF), and a 1D-CNN baseline in the non-pretrained setting. It also achieved competitive performance with SpectralFormer on pigment unmixing, despite operating at a significantly lower resolution. This indicates that the primary performance gains stem from modality-aware tokenization rather than encoder-level refinements.

We additionally studied self-supervised pretraining as a complementary component for XRF modeling. MSM improved downstream performance for both tasks, suggesting richer and more transferable spectral representations. Furthermore, the addition of the PPP objective provided additional gains, supporting its role as a task-aligned regularizer that enhances both instance discrimination and abundance estimation. While these results are promising, the current scarcity of large-scale benchmark datasets and diverse real-world samples in the CH field limits our ability to conduct more in-depth ablations or comparisons across multiple independent datasets. We hope that this work will motivate the CH community to work toward open, standardized benchmarks that will allow for the rigorous performance evaluation and comparison of future deep learning models like XRFormer. Future work will explore improved multi-task pretraining recipes for combining generic and physics-informed objectives, as well as extensions to multimodal CH pipelines integrating XRF with hyperspectral imaging.

\bibliographystyle{splncs04}
\bibliography{ref}
\end{document}